\documentclass[sigconf,nonacm, screen]{acmart}

\AtBeginDocument{%
}

\usepackage{appendix}
\begin{document}

\title{Valuing an Engagement Surface using a Large Scale Dynamic Causal Model}


\author{Abhimanyu Mukerji}
\authornote{Both authors contributed equally to this work.}
\email{abmk@amazon.com}
\orcid{0009-0005-4298-3548}
\affiliation{%
  \institution{Amazon}
  \city{Sunnyvale}
  \state{CA}
  \country{USA}
}

\author{Sushant More}
\authornotemark[1]
\email{morsusha@amazon.com}
\orcid{0000-0002-3746-2431}
\affiliation{%
  \institution{Amazon}
  \city{Seattle}
  \state{WA}
  \country{USA}
}

\author{Ashwin Viswanathan Kannan}
\email{ashvkann@amazon.com}
\orcid{0000-0002-2929-4263}
\affiliation{%
  \institution{Amazon}
  \city{Sunnyvale}
  \state{CA}
  \country{USA}
}

\author{Lakshmi Ravi}
\email{lrravi@amazon.com}
\orcid{0009-0009-4562-1234}
\affiliation{%
  \institution{Amazon}  
 \city{Sunnyvale}
  \state{CA}
  \country{USA}
}

\author{Hua Chen}
\email{huaxchen@amazon.com}
\orcid{0009-0005-2904-1513}
\affiliation{%
\institution{Amazon}  
   \city{Sunnyvale}
  \state{CA}
  \country{USA}
}

\author{Naman Kohli}
\email{namkohli@amazon.com}
\orcid{0000-0002-1896-4348}
\affiliation{%
  \institution{Amazon}
  \city{Vancouver}
  \state{BC}
  \country{Canada}
}

\author{Chris Khawand}
\email{khawandc@amazon.com}
\orcid{0009-0000-5283-9391}
\affiliation{%
  \institution{Amazon}
  \city{Seattle}
  \state{WA}
  \country{USA}
}

\author{Dinesh Mandalapu}
\email{mandalap@amazon.com}
\orcid{0009-0007-2984-859X}
\affiliation{%
  \institution{Amazon}
  \city{Seattle}
  \state{WA}
  \country{USA}
}

\renewcommand{\shortauthors}{Mukerji et al.}

\begin{abstract}
With recent rapid growth in online shopping, AI-powered Engagement Surfaces (ES) have become ubiquitous across retail services. These engagement surfaces perform an increasing range of functions, including recommending new products for purchase, reminding customers of their orders and providing delivery notifications. Understanding the causal effect of engagement surfaces on value driven for customers and businesses remains an open scientific question. In this paper, we develop a dynamic causal model at scale to disentangle value attributable to an ES, and to assess its effectiveness. We demonstrate the application of this model to inform business decision-making by understanding returns on investment in the ES, and identifying product lines and features where the ES adds the most value. 
\end{abstract}




\maketitle

\section{Introduction}
\label{sec:introduction}

Causal frameworks are a powerful tool to provide actionable insights into complex business problems. Businesses can make data-driven decisions by understanding cause-and-effect
relationships to maximize efficiency and allocate resources better. There has been an increased effort in both academic research and industry to use causal inference for informed decision-making and to measure economic impact. This work has led to the development of open source packages such CausalML \cite{CausalML} for uplift modeling
in marketing campaigns, PyWhy \cite{PyWhy1, PyWhy2} for causal reasoning and evaluating interventions, and EconML \cite{EconML1, EconML2, EconML3} for estimating heterogeneous treatment effects in customer segments.
Large organizations require accurate measurement of long-term causal impacts to calculate the returns on their investments correctly. Reliable and scalable strategies to do this are therefore increasingly important.

The share of online shopping has steadily grown around the world. In the US, the percentage of online shopping has grown from less than 1\% in 2000 to around 15\% in 2023 \cite{Pew_research_center}. Providing the best shopping experience for online customers is therefore a high priority for retail businesses around the world. Combined with developments in artificial intelligence (AI), this has led to the creation of Engagement Surfaces (ES). An ES is any application, feature, widget (or collection thereof) that is intended to help customers save time, reduce shopping effort, and support new product discovery. 

Almost all prominent e-commerce companies have sophisticated ES to improve customer shopping experiences. Additionally, there are some companies that offer integration of an engagement surface as a service to online sellers \cite{Tidio}. Depending on the level of sophistication, an ES typically performs one or more of the following functions. 
\begin{itemize}
\item 
Help customers discover new products or products that they may be interested in
\item Generate shopping lists
\item
Provide discount codes and special offers
\item
Recover abandoned shopping carts
\item
Offer personalized product recommendations and reorder suggestions
\item
Answer customer questions
\item
Track orders and returns
\end{itemize}
Despite the prevalence and sophistication of ES, it is unclear exactly how much this has enhanced customer experience, and how that has translated into revenue for the business. The economic rationale for ES hinges primarily on the use of an ES leading to increased customer spend and interactions with monetizable features in the medium to long run. However, \emph{causally} attributing increase in sales to use of the ES is a hard scientific problem given the interconnected nature of hundreds of types of customer behavior over time, as well as the breadth of concurrent features that customers are exposed to. Nevertheless, understanding this causal relationship is important to businesses as it guides the capital investment made to offer ES to customers, which may be sizable.

We also note that conducting A/B tests \cite{AB} (i.e., randomized control trials, which are generally regarded as the gold standard for causal analysis) with the full ES experience is often impossible due to practical or ethical constraints. It might be possible to use A/B experimentation for valuing a specific feature of an ES, but such an approach usually only operates with a small customer sample and requires a disciplined methodology to map to the entire population. Moreover, A/B testing may prove expensive when done repeatedly for new feature rollout, and may not be possible for all available features. This motivates our approach, where we leverage observational causal modeling, which requires additional assumptions but provides a scalable and repeatable solution paradigm. In this work, we build a large-scale Dynamic Causal Model (DCM) trained on billions of customer actions, and apply it to the problem of estimating the value generated by an ES. The DCM helps us answer a counterfactual question -- "How much money would the business lose if the Engagement Surface didn't exist at all (or didn't exist during a specific period of time)?". Measuring this associated treatment effect is a good proxy for the value generated by the ES. We highlight that while our solution is applicable to any ES, we focus in particular on cases where ES features may be highly interconnected in customer usage patterns, and may drive the most value when operating synchronously. 

To the best of our knowledge, this is the first paper presenting a scalable and generalizable causal model for estimating the value generated by an engagement surface. The paper is structured as follows. In Sec.~\ref{sec:DCM}, we provide an introduction and motivation for the dynamic causal model. In Sec.~\ref{sec:contemp_effects}, we discuss inclusion of contemporaneous treatment effects arising from concurrent customer behavior (which is relevant for the use case of valuing an ES). We provide an overview of implementation in Sec.~\ref{sec:implementation} and showcase results in Sec.~\ref{sec:results}. We end with conclusions and our recommendations for further study in Sec.~\ref{sec:conclusion}.     

\section{Dynamic Causal Model}
\label{sec:DCM}

We developed a Dynamic Causal Model (DCM) to identify mediated causal effects and interactions in high dimensions over a long period of time in order to generate interesting aggregate business counterfactual analyses. One such example would be estimating the overall causal effect of removal of specific product lines or features. This model was developed in part to avoid some of the limitations of binary treatment effect estimation of a single customer action \cite{More_DML}. The DCM models the interaction across time periods between all model variables and outcomes of interest (typically customer spending or revenue generated). We refer to variables capturing customer interactions with the shopping interface as surrogates --- following terminology in econometrics for short-term observed features associated with longer-term outcomes of interest. These are typically structured to identify customer behavior such as purchases and new program sign ups. The dynamic framework offers numerous advantages, notably by allowing treatment effects in one period to impact subsequent surrogates, generating cascading spillover effects, and leading to mediated causal pathways that are important to account for. Moreover, such a structure also enables joint-policy evaluation, where we can consider complex counterfactual scenarios where a large group of causal pathways may be blocked off, attenuated or enhanced. Finally, we resolve the issue of evaluating the combined impact of different surrogates, which could be double counted or incorrectly aggregated if being studied through separate binary interventions, as is usually the case in online A/B testing.

The DCM framework models hundreds of customer surrogates/actions over time and parametrizes the associations between them.
\begin{figure}[h!]
  \centering
  \includegraphics[width=0.85\linewidth]{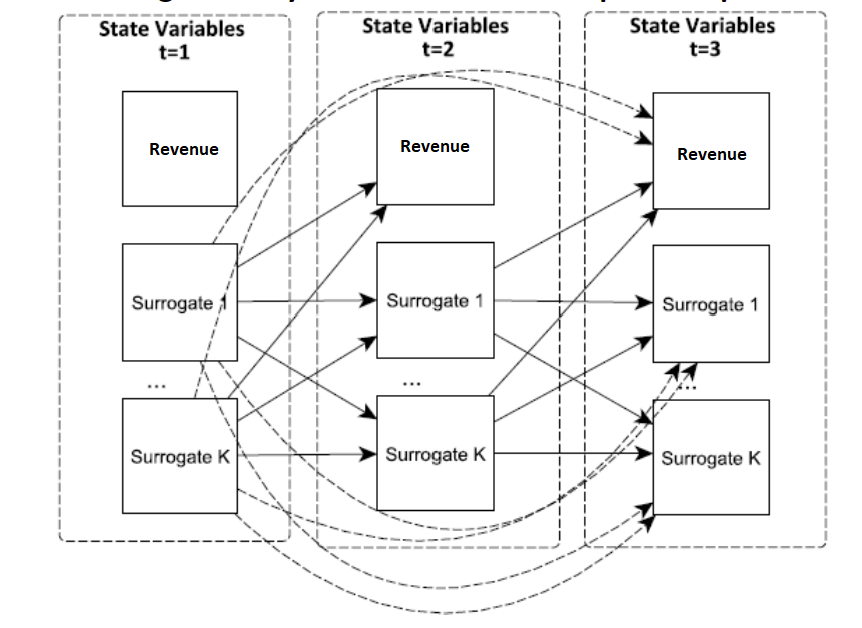}
  \caption{DCM Causal Graph: 3-period example. The solid arrows represent direct relationships. The dashed arrow represent indirect (mediated) relationships.}
  \label{fig:dcm_causal_graph}
\end{figure}
Figure~\ref{fig:dcm_causal_graph} shows a simplified 3-period example to explain the key elements of the model. We measure the relationship between surrogate $1$ and surrogate $K$ from $t=1$ to $t=2$ (solid arrow), but can also observe the relationship between surrogate $1$ and surrogate $K$ from $t=1$ to $t=3$ (dashed arrow), and vice versa between surrogate $K$ and surrogate $1$. The surrogate variables help isolate the possible channels through which a causal effect propagates over time to outcomes or other surrogates. This allows us to quantify how surrogates at a particular point in time affect other surrogates or outcomes in future periods. A major advantage of such a model is that we properly capture mediated effects, where changes in a surrogate at a point in time continue to generate ripple effects in future periods through interactions with subsequent surrogate variables.

The model estimates a set of equations which capture surrogate interactions as described above. The solution strategy we employ estimates these associations and then simulates counterfactuals that capture the state of the ecosystem at any point in time (or across time) when specific input surrogates are `shocked'. A `shock' here refers to a scenario where a surrogate (or a group of surrogates) is increased/decreased (or set to zero) for a point in time, or over a period of time. The model propagates this shock through the dynamic system of equations and then aggregates the effects to generate a counterfactual (i.e., under the shocked condition) measure that can be used to compute outcomes of interest across a chosen period. Surrogates can be configured to represent product lines or entire businesses, and the model can therefore capture the effect of `shutting down' this group of customer behaviors. The DCM is an improvement over the causal impact estimation framework for a single customer action \cite{More_DML}. It is able to handle overlapping interactions amongst customer events and provides a way to estimate long term causal effects in a scientifically disciplined way.  

\subsection{DCM Training}

While non-linear versions of the DCM are possible to construct, we define for this application a partially linear model to evaluate an outcome $y_t$ with respect to binary policy $D$, and define $S$ to be the surrogate variable which captures the impact of any exogenous policy change attributed to $D$. $f(.)$ and $g(.)$ represent functions controlling for variation attributable to pre-determined customer features $X_0$. A customer’s surrogate and revenue values at time $t$ are defined by
\begin{eqnarray}
S_t &=&  \sum_{s=0}^{t-1} S_s \pi_{t,s} + f_t(X_0) + D \, \theta_t+  \nu_t \nonumber \\ 
y_t &=&  \sum_{s=0}^{t-1} S_s \beta_{t,s} + g_t(X_0) + D \, \eta_t+  \epsilon_t
\label{eq:standard_dcm_eqs}
\end{eqnarray}
where $\pi_{t,s}$, $\beta_{t,s}$, $\theta_t$, $\eta_t$ are the coefficients to be learned during model training. $\nu_t$ and $\epsilon_t$ are zero-mean noise terms. 

Note that Eqs.~\eqref{eq:standard_dcm_eqs} are modeled per customer $i$. We omit the subscript $i$ for simplicity of exposition. The DCM framework internally uses a scalable regression algorithm to estimate hundreds of regressions in Apache Spark \cite{Spark}. It works with a JSON input configuration which keeps it modular and flexible. Importantly, we model customer actions and outcomes at a given time period as a function of that customer's lagged actions. That is, the baseline model uses only past behavior on the right hand side of the regression. The reason for this is that the model consumes data in a panel format, and since we do not use timestamp-level information, we cannot disentangle the causal direction of actions that occur in the same time period. As we discuss in Section \ref{sec:contemp_effects}, this can neglect intra-period effects that could be relevant for modeling ES --- we overcome this by imposing some additional restrictions on the causal graph implied by the DCM.


\subsection{DCM Scoring}
\label{sec:dcm_scoring}

The second aspect of the framework involves the simulation tool that uses the trained model parameters to predict potential outcomes under alternate scenarios. Counterfactuals for business or product line elimination scenarios are computed by defining shocks that change the outcomes of interest (e.g. customer spending/ unit sales) and surrogates that belong to that specific business by a percentage amount. These shocks are applied in a sequential manner because of which the dynamic structure cannot be solved in the general case without recursion. For example, we simulate counterfactuals in the above setup of Eq.~\eqref{eq:standard_dcm_eqs} as follows: 
\begin{enumerate}
\item
Define or estimate intervention $\theta_t$ (e.g., policy shock, surrogate size change).
\item
$t=0$: Compute $\{ S_0^{(1)}, S_0^{(0)}\}$ directly off the inputs $\{X_0, \nu_0, \theta_0\}$. This represents features at time $t=0$.
\item 
$t=1$: Compute $\{ S_1^{(1)}, S_1^{(0)}\}$ from $\{ S_0^{(1)}, S_0^{(0)}\}$ and predetermined inputs $\{X_0, \nu_1, \theta_1\}$. $\theta_1$ is the shock generated and the computed counterfactuals at $t=1$ are used as input for the next time step.
\item 
$t=2$: Compute $\{ S_2^{(1)}, S_2^{(0)}\}$ from $\{ S_1^{(1)}, S_1^{(0)}\}$ and predetermined inputs $\{X_0, \nu_2, \theta_2\}$.
\item
Repeat this process until you have reached the total number of time periods $N$. 
\end{enumerate}

\section{Contemporaneous treatment effects}

\label{sec:contemp_effects}

In this section we will discuss an enhancement to the standard DCM model (Eq.~\eqref{eq:standard_dcm_eqs}) which is relevant when we use the model to estimate value driven by an ES.  

\subsection{Business Motivation}

\label{sec:buss_motivatn}

Revisiting Fig.~\ref{fig:dcm_causal_graph}, we note that the standard DCM framework captures lagged causal effects (cause preceding the effect) on future surrogates and outcomes at least one time unit apart. In practice, the DCM framework is implemented using a weekly time grain. That is, we aggregate  activity for each customer on a weekly basis and model Eqs.~\eqref{eq:standard_dcm_eqs}. Returning to the business use case of ES, we note that an important value-add arises from friction reduction for customers making purchases. As a result, value driven by the ES is likely associated with near-simultaneous customer actions (e.g., when a customer engages with the ES to obtain information about an item, adds it to cart, and completes their purchase in the same hour). Since the DCM operates using data at a coarse time grain, we refer to such actions as “contemporaneous” or “same-period” actions. The DCM enhancement discussed here is important as it ensures that we properly attribute customer actions in the same period, mediated through the ES, towards its valuation. Note, given the DCM structure, we regard any customer actions in that same time period (e.g., week) as being `simultaneous' from a causal perspective.

\begin{figure*}[h!]
  \centering
  \includegraphics[width=\linewidth]{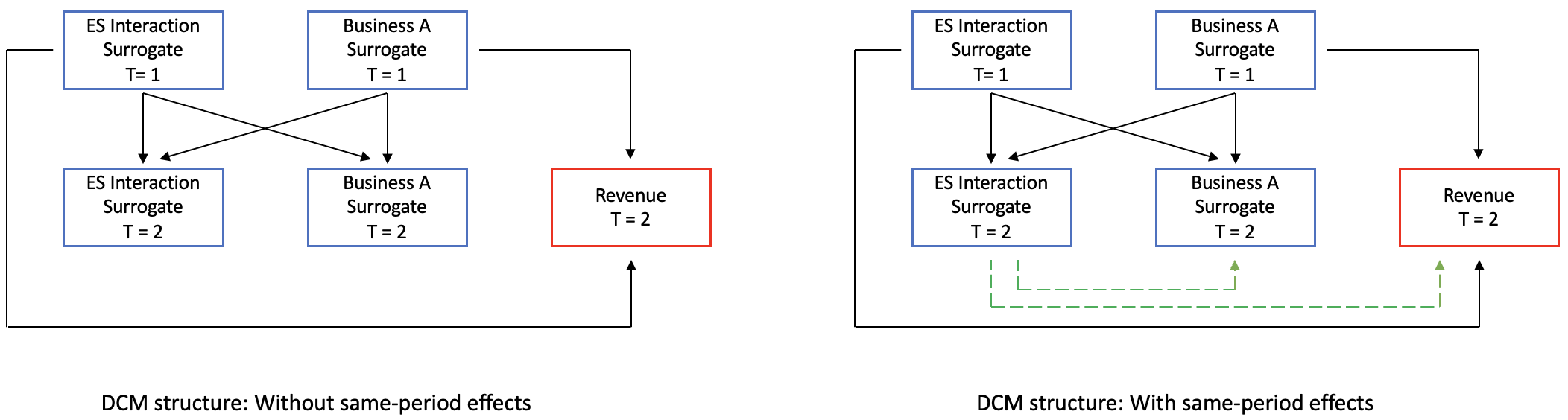}
  \caption{Simplified representation of the DCM causal graph with and without the same-period effects.}
  \label{fig:dcm_causal_graph_with_wo_same_period}
\end{figure*}

To explain the intuition behind the same-period effects module in our DCM, we consider a simple case with two businesses, the ES, and business A\footnote{placeholder for any business which is not directly related to the ES under consideration}, and two time periods. Both businesses are assumed to be represented by a single surrogate, and the outcome of interest (revenue in this example) is realized in the second period (with T=1 as a “pre-period” of lagged surrogates for illustration). The causal graph corresponding to this conventional/standard causal relationship in DCM is shown in the left panel in Fig.~\ref{fig:dcm_causal_graph_with_wo_same_period}.

If we modify this structure to incorporate same-period effects, our model now captures how events in a particular time period produce causal impact on other events and outcomes in that same time period. This is designed to account for the near-simultaneous cause and effect relationships such as those posited for an ES. In this example with same-period effects, any near-simultaneous (and therefore within one week/day/our minimum time grain) interactions with the ES that lead to a purchase transaction are attributed to have been caused by the ES. Returning to the causal graph in Fig.~\ref{fig:dcm_causal_graph_with_wo_same_period}, we allow for additional causal links (represented by green dashed arrows) in the panel on the right. This enables the model to learn associations between outcomes/surrogates of interest and ES interactions in that same time period. We note that this requires specifying the direction of the causal link as an identification assumption, which in the case of the ES is assumed to run from the ES surrogate – this is a necessity given the fidelity available with this time grain. That is, in the case of near-simultaneous events, we give the ES credit for driving the causal effect. This is reasonable in the case of an ES driving a revenue outcome, but would lead to misspecification if ES interactions are in fact driven by actions/surrogates in other businesses in the same period. Given that the ES functions as a friction reducer for customers engaging in purchase behavior, we expect this latter case is relatively infrequent with muted bias induced in measured treatment effects as a result. Despite these potential limitations, we choose to account for same-period effects within the DCM framework to allow for ease of use and repeated analyses, rather than using an auxiliary model to obtain contemporaneous incrementality inputs. One of our ongoing enhancements to this framework relies on using short term A/B testing to learn the direction of causal links and adjusting the regression structure accordingly. 

In the next section, we provide the mathematical details of the framework to include the contemporaneous effects in the DCM.

\subsection{Mathematical Framework}

Let us denote $S$ as the complete set of surrogates. Let $I^A$ be the surrogates specifying the interactions with the ES and $S^{NA}$ be the non-ES surrogates. 
So, $S = S^{NA} \cup I^A$ and $S^{NA} \cap I^A = \{\}$.

Based on the business context discussed in Sec.~\ref{sec:buss_motivatn} and taking motivation from the literature discussing vector autoregression models \cite{Var1, Var2, Var3, Var4}, we update the standard DCM equations from Eq.~\eqref{eq:standard_dcm_eqs} as follows:
\begin{eqnarray}
y_t &=& \delta \, I^A_t +  \sum_{i=0}^{t-1} I_i^{A} \kappa_{t,i}^{A} + \sum_{k=0}^{t-1} S_k^{NA} \kappa_{t,k}^{NA} +  q_t(X_0) + \psi_t \nonumber \\
S_t^{NA} &=& \gamma \, I^{A}_t + \sum_{i=0}^{t-1} I_i^{A} \beta_{t,i}^{A} + \sum_{k=0}^{t-1} S_k^{NA} \beta_{t,k}^{NA} + g_t(X_0) + u_t \nonumber \\
I_t^{A} &=& \sum_{i=0}^{t-1} I_i^{A} \rho_{t,i}^{A} + \sum_{k=0}^{t-1} S_k^{NA} \rho_{t,k}^{NA}  + m_t(X_0) + \upsilon_t
\label{eq:same_period_eqns_expanded}
\end{eqnarray}

We can simplify the right-hand side (RHS) of Eqs.~\eqref{eq:same_period_eqns_expanded} to use $S$:
\begin{eqnarray}
y_t &=& \delta \,I^A_t +  \sum_{i=0}^{t-1} S_i \, \kappa_{t,i}  +  q_t(X_0) + \psi_t 
\label{eq:yt_same_period} \\
S_t^{NA} &=& \gamma \, I^{A}_t + \sum_{i=0}^{t-1} S_i \, \beta_{t,i}  + g_t(X_0) + u_t \label{eq:st_same_period} \\
I_t^{A} &=& \sum_{i=0}^{t-1} S_i \, \rho_{t,i} + m_t(X_0) + \upsilon_t
\label{eq:It_same_period}
\end{eqnarray}

Recall that the Eq.~\eqref{eq:same_period_eqns_expanded}-Eq.~\eqref{eq:It_same_period} are modeled at the customer level. We have dropped the customer index here for brevity. 

As pointed out in the Sec.~\ref{sec:buss_motivatn}, business context is necessary to specify the appropriate causal links in the same-period equations. In Eqs.~\eqref{eq:yt_same_period}-\eqref{eq:It_same_period}, we assume following constraints on the causal graph:
\begin{itemize}
\item
Only ES interactions can impact surrogates/outcomes in the same period. This is denoted by the presence of the $I_t^A$ term on the RHS of Eqs.~\eqref{eq:yt_same_period} and \eqref{eq:st_same_period}. 
\item
Non-ES surrogates cannot cause any surrogate/outcome in the same period. This is the standard DCM temporal constraint to avoid causal loops. 
\item
ES interactions of one type cannot cause ES interactions of another type in the same period. This is indicated by equation for $I_t^A$ only having the lagged surrogates on the RHS. 
\end{itemize}

To generalize to other business cases, we would allow same period causal effects from the business of interest onto any surrogate/outcome subject to  restrictions based on business knowledge of the direction of causality (eg. actions cause outcomes but not vice versa). We are currently working on improving this through causal graph discovery in the same period, that would allow the model to learn the appropriate causal directionality and impose these constraints in the regression structure.

In the next section we give an overview of the implementation of the same-period DCM model for the ES use case.

\section{Implementation overview}
\label{sec:implementation}

The high-level implementation steps are as follows:
\begin{enumerate}
\item
We analyze the weekly spending of millions of customers for 3 years (denoted by $y_t$ in Eq.~\eqref{eq:yt_same_period}). 
\item 
We also analyze the weekly participation of the same sample of customers in the standard surrogates (denoted by $S_{NA}$ in Eq.~\eqref{eq:same_period_eqns_expanded}). We have about 2000 such surrogates today to represent customer behavior. 
\item
To the above two, we join the weekly interactions of the customers with the ES. 
As mentioned in Sec.~\ref{sec:introduction}, customers interact with the ES in multiple ways. We worked with business partners to consolidate the different customer interactions into feature groups.  
These are denoted by $I_t^A$ in Eq.~\eqref{eq:same_period_eqns_expanded}. The ES we are working with operates in three broad channels, and we analyze the customer participation in ES feature groups of interest across all these three channels. 
\item
Based on the historical surrogate/ interaction feature participation data and the historical spend, we learn the coefficients in Eqs.~\eqref{eq:yt_same_period}-\eqref{eq:It_same_period}. 
\item 
Once the model has been trained, to estimate the value of a certain ES channel or feature we use the model to construct the counterfactual scenario where the specified ES channel/feature is unavailable to customers. This is done through dynamic model scoring as described in Sec.~\ref{sec:dcm_scoring}. 
\end{enumerate}

In Appendix~\ref{sec:implementation_details}, we discuss some nuances related to DCM training and scoring while incorporating same-period effects. We note that inclusion of same-period effects has no impact on computational efficiency or model run time. 

\subsection{Shapley Attribution}

The DCM framework also allows us to apply attribution algorithms to business valuations. This becomes relevant when estimating the model for multiple businesses that jointly generate value to ensure that value is not being double counted. An important application of such valuations for business stakeholders is in sizing financial investments and optimizing product selection. To do this effectively, we needed to develop the capability to structure model output for direct comparison with financial accounts. In other words, valuations produced by the DCM framework should tally with revenue reported in financial statements. Shapley Attribution \cite{Shapley1, Shapley2} is our default method of doing this, and is effectively based on a weighted average of how much a business contributes when it is added to a
group of other active businesses at the margin, across all combinations. In the case of ES valuation, this allows us to understand the marginal contribution to different product lines given the investment required to support ES tooling. More details regarding this are provided in Appendix~\ref{sec:Shapley}.

\subsection{System Architecture Overview}

\begin{figure*}[h!]
  \centering
  \includegraphics[width=0.7\linewidth]{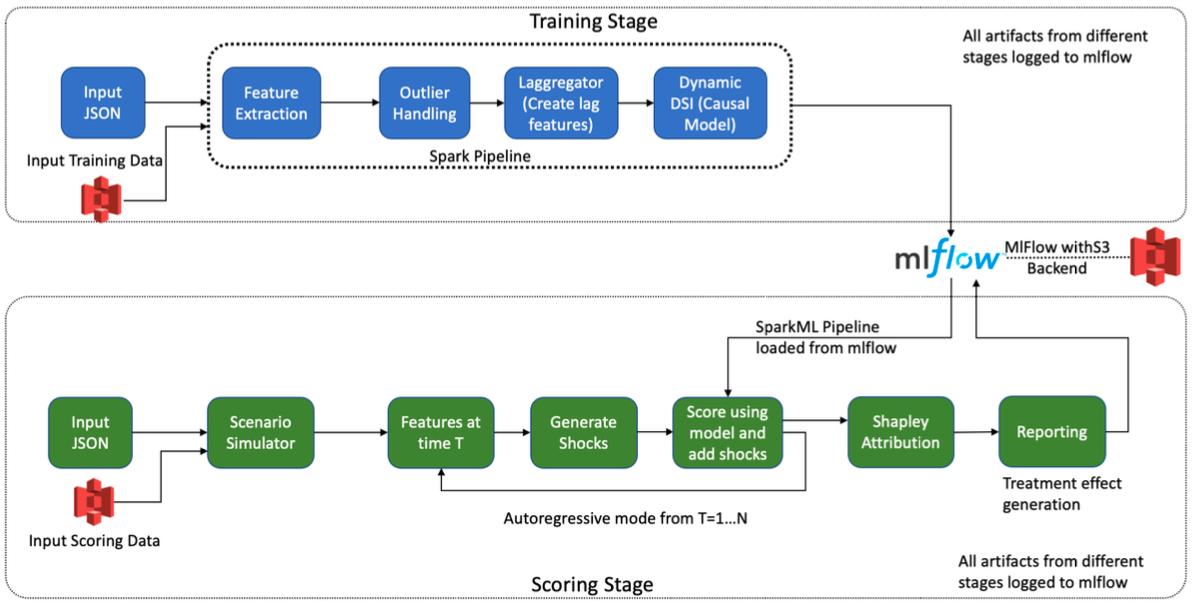}
  \caption{DCM framework architecture with training and scoring stages.}
  \label{fig:architecture_overview}
\end{figure*}
Figure~\ref{fig:architecture_overview} showcases the current system architecture of the DCM framework. The framework has been developed using modular components in order to prioritize ease of maintenance and scalability. There is minimal dependency between modules enabling faster, parallel development cycles and reducing downtime if certain components need to be updated. We emphasize adaptability to a wide range of use cases, and interpretability (transparent insights that help us understand what is driving any result). 
The model is implemented in Apache Spark, and is structured with a pipeline design pattern that processes data in a series of stages, with each stage performing specific operations on the data. We build custom SparkML pipelines using Transformers and Estimators in Spark as required. An added advantage is that the inherited serialization and deserialization persist when loading the fitted models. This design pattern avoids any replication of code in training/scoring orchestration or manual configuration changes to map with trained models.

The DCM framework utilizes a stateless design function where model training and scoring are designed to be independent stages. The scoring input can be processed by using a training pipeline without requiring any further context or other state information. This simplifies the parameters needed during inference and enables easy broadcast of functions/parameters to enable distributed scoring. 

Experimentation at high velocity is another key feature of the framework. We achieved this by creating batch tooling that is powered through AWS Lambda \cite{AWSlambda} to execute multiple parallel inference runs. The base model configuration can be overridden in parallel runs that spin up and terminate clusters after completion. All artifacts are logged into MLFlow \cite{mlflow} enabling easy tracking of experiments, reproduction of model results, and sharing model runs across different users. 
The linear model structure enables us to disentangle the effect of each surrogate variable on outcomes and other surrogates, and recovering coefficient values allows us to interpret causal effects and provide insights on customer behavior. The DCM model uses the MLFlow model registry to log models that can be read back to examine the learned weights. Additionally, the framework allows for confidence interval estimation on valuations through bootstrapped runs.
The ES model runs use $\sim$100 million customers analyzed over 3 years at a weekly grain, with over 2000 features per time period.

This has been possible using PySpark and SparkML \cite{SparkML} Transformers and Estimators and utilizing sparse vectors for data compression. Another optimization that we added is in-memory creation of lagged features used on the RHS of Eq.~\eqref{eq:standard_dcm_eqs}.
We score the model linearly across time periods, using the input scoring data and the trained model pipeline. This iterative autoregressive approach is suboptimal given Spark's lazy computation, which relies on creating dynamic plans first and then executing the computation graph. This leads to slow execution time, as the computation graph becomes intractably huge. To overcome this, we utilized checkpointing to break the lineage in the computation graph and force periodic calculation.

\section{Results}
\label{sec:results}

In this section we discuss our main results, estimating the value generated by the ES using the DCM framework. 
We provide valuations for individual ES channels as well as the overall valuation. We anonymize the channel names to preserve business confidentiality. Note that the total ES value is not a simple sum of individual channels because some of the value of each channel is jointly determined with other channels (i.e., the incremental value of each channel is affected by the presence of other channels). 

The ES generates value by improving customer satisfaction leading to increased engagement and purchases. Along with the total incremental spend across all channels (which corresponds to the total ES value), we also look at different slices across the three ES channels and the different product groups that ES facilitates shopping for. 
Numbers in Table~\ref{tab:dcm_ops_with_same_period} answer questions like “How much revenue would have been lost in each of the product groups in the full year 2022 if the ES was not available in a certain channel?”
Product groups are broad categories such as grocery, electronics, apparel etc. Looking at the value for different product groups allows us to gain insights into whether the ES adds more value in one category versus another. To maintain business confidentiality, we scale all numbers in Table~\ref{tab:dcm_ops_with_same_period} through Table~\ref{tab:dcm_ops_sa_interaction_features_attributable_to_same_period} between 0 and 100 by normalizing with the total estimated causal value for the ES. Additionally, we also anonymize the product group names, channels and features. The key insights from our results are as follows:
\paragraph{1. Top channel drives over 50\% of ES value:}
Based on Table~\ref{tab:dcm_ops_with_same_period}, we see that the highest monetary value for ES comes from Channel 2. 
This is consistent with the idea that this was a channel with broader reach driving the most customer engagement and hence the most monetary value. 
We also produced bootstrap confidence bounds to quantify sampling error. These bounds are generally quite
narrow, even at the product group level (typically within 5-7\% of the point estimate). 
\begin{table*}[h!]
\caption{FY 2022 normalized incremental revenue across Product Group facilitated by the ES. This is the value causally attributed to ES by the DCM framework.}
\begin{center}
\includegraphics[width=0.7\linewidth]{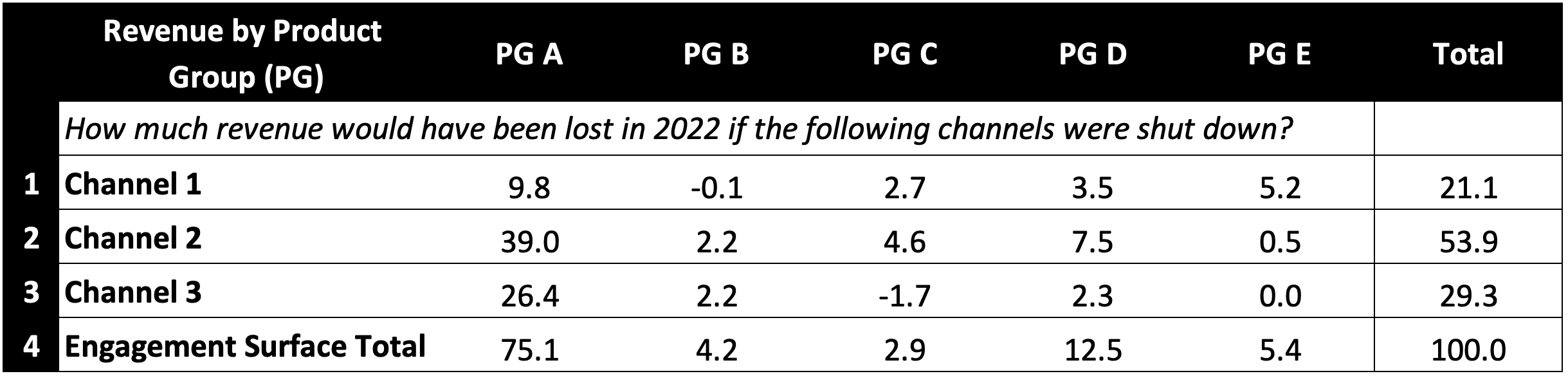}
\end{center}
\label{tab:dcm_ops_with_same_period}
\end{table*}
\begin{table}[h!]
\caption{FY 2022 normalized incremental revenue for the top ES interaction features groups}
\begin{center}
\includegraphics[width=0.9\linewidth]{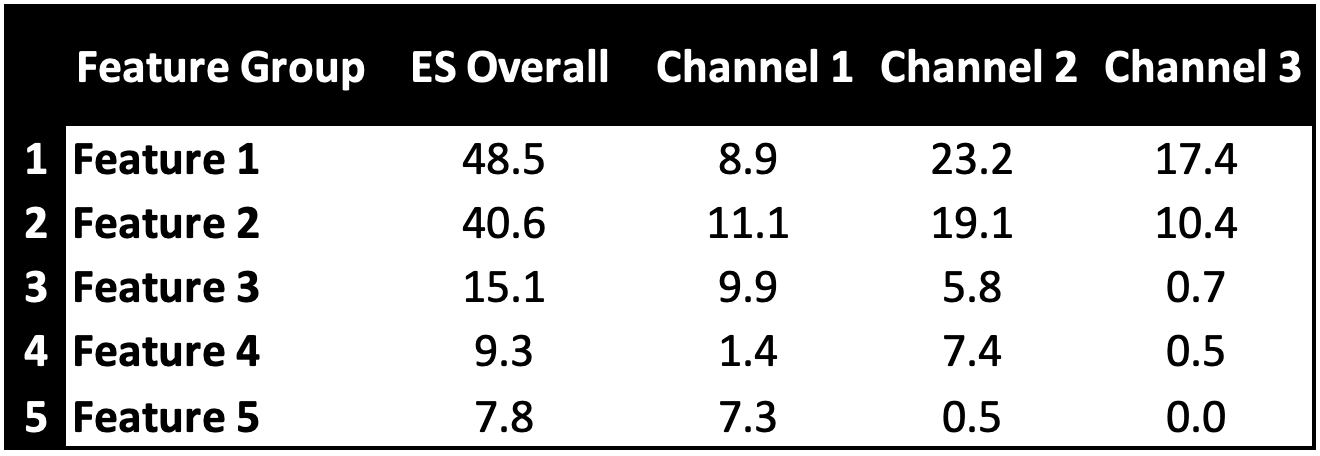}
\end{center}
\label{tab:dcm_ops_sa_interaction_features}
\end{table}
\paragraph{2. The ES drives the highest value in low average selling price (ASP) and regular purchase categories:} When looking at the incremental impact of ES interactions on the different product group categories, we notice that over 85\% of them are from low ASP and frequent purchase categories. Examples includes categories such as baby products, beauty products, apparel, and grocery. Customers may be more interested in engaging with the ES for products that are more regularly purchased, where Q\&A (to gather product information) and price checks are more useful, and where a customer’s time investment in product selection and search is less beneficial. Our findings are overall consistent with our intuition given the nature of ES features: friction-reduction customer experience (in the current form available) drives the most value in product groups that customers are likely
already familiar with, or that are more frequent and/or low ASP buys for which customers perceive less risk in acting based on
purchase-relevant information provided through the ES.
\paragraph{3. Top two features drive ~73\% of ES's impact:}
We also estimated the value lost if specific ES interaction features (such as those offering tracking or product information) were not available.
Table~\ref{tab:dcm_ops_sa_interaction_features} shows valuations for each interaction feature for the ES as a whole, as well as broken out by the different
channels. The effects in this table are not strictly additive, but still helps us identify the major value driving features. We observe that the bulk of the value (over 73\%) comes from the top two feature categories. These feature categories are associated with re-buys and finding product information. Given that most of the ES value is driven by low-ASP/ regularly purchased products, it is reasonable that features supporting efficient repeat orders and product information provision are important for the ES.
\paragraph{4. Near-simultaneity of ES interaction and purchases plays a large role in value driven:}
As mentioned in Sec.~\ref{sec:contemp_effects}, we updated the standard DCM model to better capture causal flows between the ES and other customer actions in the same time period (one week in the model specification we used).   
The output from the model which does not include the same-period effects is shown in Table~\ref{tab:dcm_ops_wo_same_period}. For ease of interpretation, Table~\ref{tab:dcm_ops_results_attr_to_same_period} shows the difference in incremental revenue between Table~\ref{tab:dcm_ops_with_same_period} and Table~\ref{tab:dcm_ops_wo_same_period}. This captures the value that we would have missed had we not properly accounted for same-period effects in the DCM.  
\begin{table*}[h!]
\caption{FY 2022 normalized incremental revenue by Product Group excluding the same-period effects.}
\begin{center}
\includegraphics[width=0.7\linewidth]{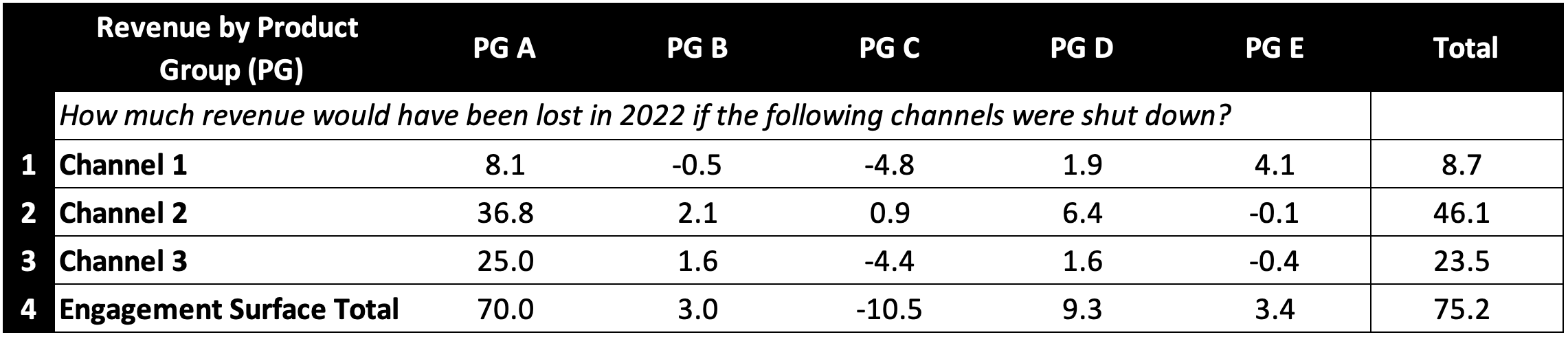}
\end{center}
\label{tab:dcm_ops_wo_same_period}
\end{table*}
\begin{table*}[h!]
\caption{FY 2022 normalized incremental revenue by Product Group attributable to the same-period effects.}
\begin{center}
\includegraphics[width=0.7\linewidth]{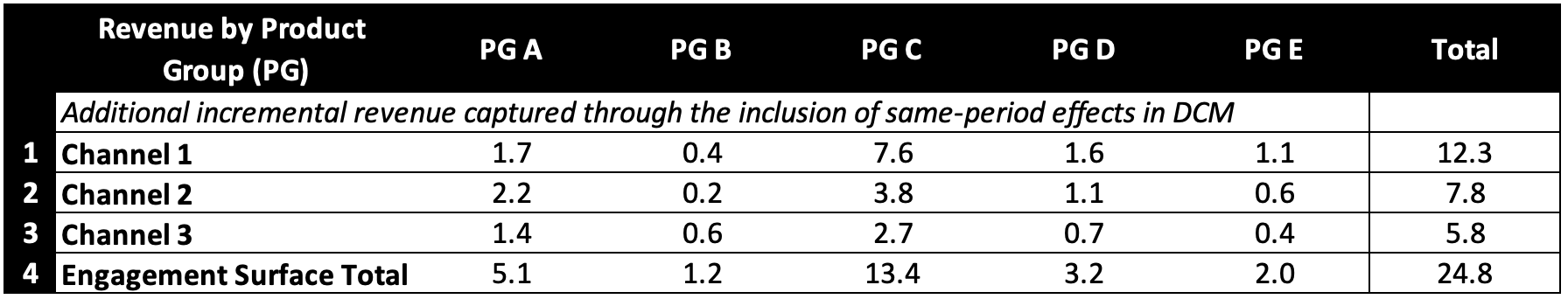}
\end{center}
\label{tab:dcm_ops_results_attr_to_same_period}
\end{table*}
We were previously concerned that omitting these same-period effects would result in undervaluation of the ES. Indeed, when we disable same-period effects, we estimate that the value of the ES is 25\% smaller. Comparing the three channels, the impact of same-period effects is again intuitive. The impact is larger in channels where the customers are most likely to complete purchases within a shorter time frame.
\begin{table}[h!]
\caption{FY 2022 normalized incremental revenue for the top ES interaction features attributable to the same-period effects}
\begin{center}
\includegraphics[width=\linewidth]{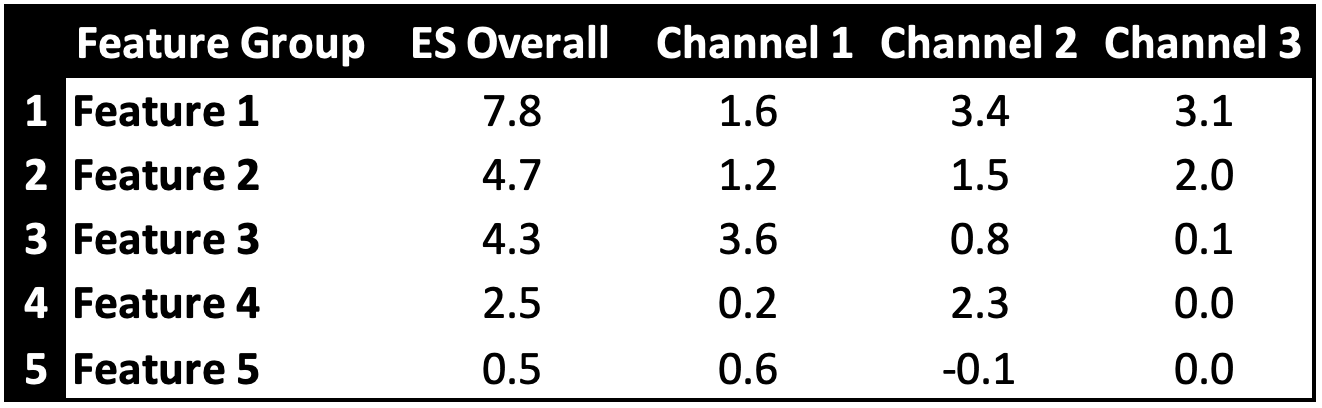}
\end{center}
\label{tab:dcm_ops_sa_interaction_features_attributable_to_same_period}
\end{table}
In Table~\ref{tab:dcm_ops_sa_interaction_features_attributable_to_same_period}, we  provide the contribution of same-period effects to the individual feature valuations. Comparing this to Table~\ref{tab:dcm_ops_sa_interaction_features}, we notice that for certain features over 25\% of the valuation comes from the same-period effects. This again underscores the importance of incorporating same-period effects when valuing customer-initiated upper-funnel actions such as those produced through ES interaction.

\section{Conclusion and Future Work}

\label{sec:conclusion}

In this work, we introduced a Dynamic Causal Model (DCM) which helps us identify the causal effect of a business under consideration, on value generated for the company. The strength of the model lies in its ability to extract causal insights in a fundamentally multivariate scenario with complex interdependencies. The DCM achieves this by modeling interactions between thousands of customer actions over several years. The recursive dynamic scoring approach allows us to construct interesting business counterfactuals by "shocking" (changing and projecting based on the model) the relevant outcomes and surrogates. The modular implementation using Apache Spark allows us to scale, and our pipeline architecture enables flexibility in analyzing different use cases efficiently.

We applied the DCM framework to evaluate the value generated by an Engagement Surface (ES). Companies leveraging ES hope to provide a better shopping journey for online customers. However, the only way an ES generates value is by increased spending among the customers that interact with ES. DCM provides a principled causal framework to estimate this value. Absent such a model, analysts studying the financial value proposition would run the risk of neglecting to account for selection bias and mediated effects, and to properly control for other predictive surrogates. 

We noted that since the ES acts as a friction reducer for customers making purchases, the value driven by the ES depends on near simultaneous or same-period customer actions. We discussed our improvements to the DCM framework to account for the effects of these same-period actions. We found that failing to include the same-period effects underestimates the total ES value by 25\%. The ES value is more affected by same-period effects in the channels where customers are likely to complete purchases within a shorter time frame.

Our work also sheds light on certain interesting aspects of customer behavior when it comes to engaging with an ES. We observed that customers are more interested in engaging with the ES we examine for products that are more regularly purchased, have a lower ASP, and where customer's time investment in product selection is less beneficial. We also demonstrated the versatility of our framework through its ability to obtain the value driven by individual ES features groups. Modifying the shocks allows us to construct counterfactuals of interest and thereby study the value created by the ES across arbitrary segments. 

Causal-based valuations like the ones we provided drive important long-term capital allocation decisions. They are especially challenging to obtain when there are multiple businesses within an organization and these business create value by supporting other business directly or indirectly. While we focus here on the use case of valuing an ES, we emphasize that our framework is flexible and can be leveraged for valuing any business or feature set of interest.  

 \subsection{Future Work}
 
 \subsubsection{Validation}
 
 One of the challenges in validating observational causal estimates is summarized by the \emph{Fundamental Problem of Causal Inference} \cite{FPCI}. The lack of observable ground truth makes it difficult to validate the output of a causal model. Businesses often use online RCTs for individual feature launch decisions. A naive way to obtain total value from RCTs is through rolling up estimates across different experiments. The major flaw with such bespoke summation is that RCTs are often sequential tests of individual feature improvements rather than holdouts of the entire customer experience that a program (such as the Engagement Surface in our example) offers. There are also secondary issues related to differential timing of the impact of RCTs, sequencing of treatments, and the role of long-term effects when annualizing short term experimental evidence. 

Building an apples-to-apples technique of comparing valuations from the DCM model to RCT-based valuations is part of our ongoing work.   
     
\subsubsection{Non-linear functional form}

In its current form, the DCM framework accounts for interdependencies among different customer action through a system of linear equations (e.g., Eq.~\eqref{eq:standard_dcm_eqs}). The use of linear system makes our framework easier to interpret and helps with building confidence in results through its transparency. It also speeds up the computation. We had previously noted that our methodology is agnostic to the specifics of the modeling framework. There is a case to be made that using a non-linear model will help us capture more complex patterns of customer action interdependencies and we have been exploring it. Leveraging transformer architecture \cite{attention} instead of the current lag structure is another active area of investigation.


\begin{appendices}
	\appendix

\section{Implementation details}
\label{sec:implementation_details}

In this section, we discuss the implementation details of the same-period model enhancement. 

\begin{figure}[h!]
  \centering
  \includegraphics[width=0.95\linewidth]{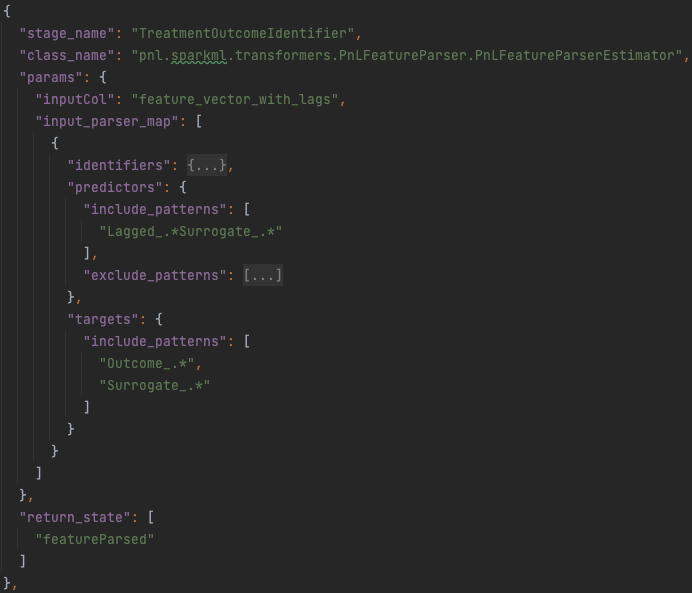}
  \caption{JSON config snippet depicting the regression structure in the standard DCM framework. }
  \label{fig:regressors_json_std_dcm}
\end{figure}

As touched upon in Sec.~\ref{sec:DCM}, the DCM framework uses a JSON input configuration. In the standard DCM, as defined in Eqs.~\eqref{eq:standard_dcm_eqs}, the outcome and surrogate variables at time $t$ are regressed on the lagged surrogates (i.e., surrogate values at the previous time steps). As the same features are used on the RHS for both outcome and surrogates, the section of the input training config that defines the regression structure is given by a sample snippet in Fig.~\ref{fig:regressors_json_std_dcm}. 

\begin{figure}[h!]
  \centering
  \includegraphics[width=0.65\linewidth]{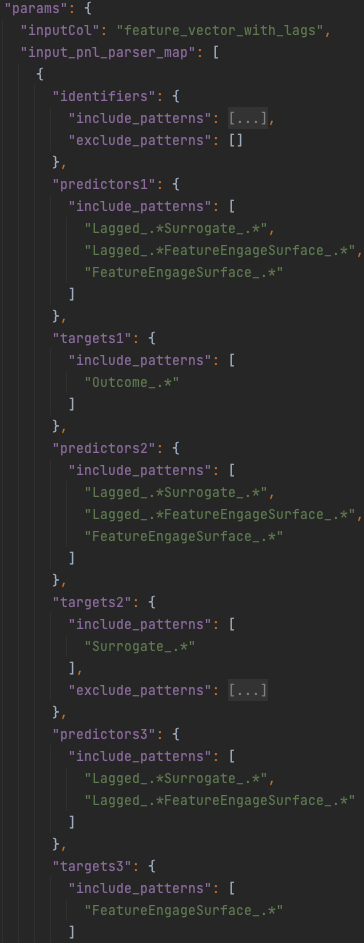}
  \caption{JSON config snippet depicting the regression structure in the DCM framework with same-period effects included.}
  \label{fig:regressors_json_same_period_dcm}
\end{figure}

The regression structure changes when we want to incorporate same-period effects as shown in Eqs.~\eqref{eq:same_period_eqns_expanded}. Specifically, the predictors are different depending on if we are estimating the ES interaction features or the non-ES surrogates (and Outcomes). This is reflected in the JSON config as show in Fig.~\ref{fig:regressors_json_same_period_dcm}

We updated the DCM training code to make sure it can work with different predictors based on the choice of target. One nuance to note during model scoring is that the value of ES interaction feature at time $t$ (denoted by $I_t^A$ in Eq.~\eqref{eq:It_same_period}) needs to estimated to obtain the value of outcome and non-ES surrogates at the same time period $t$ (denoted by $y_t$ and $S_t^{NA}$ in Eqs.~\eqref{eq:yt_same_period} and \eqref{eq:st_same_period} respectively). This means that during same-period scoring, Eq.~\eqref{eq:It_same_period} needs to be evaluated first before either Eqs.~\eqref{eq:yt_same_period} or \eqref{eq:It_same_period}. We updated the DCM scoring module to make sure this is properly accounted for.

\section{Shapley values}
\label{sec:Shapley}

Shapley attribution for business $k$ is formally specified as:
\begin{equation}
\phi_k = \dfrac{1}{\#~players} \sum_{coalitions~excluding~k} \dfrac{marginal~contr.~of~player~k}{\#~coalitions~excl.~k~of~this~size}
\end{equation}
We explain this through the 2-surrogate case, just to keep notation simple. Assume each business has one surrogate and we were interested in summarizing impact of 
manipulating each surrogate with a single value. Let $s^{*}_k$ be the realized level of the surrogate, and $s^{0}_k$ be some counterfactual baseline level. The Shapley attribution will be a weighted average of 4 differences in outcomes. Each value corresponds to the thought experiment of a business `adding’ itself to a coalition that does not include it, broken down below.
\newline
{\textbf{Benefits from increasing business 1’s surrogate value from $s^{0}_1$ to $s^{*}_1$:}}
\begin{eqnarray}
V_{1,1} \equiv F(S_1 = s^{*}_1, S_2 = s^{*}_2) - F(S_1 = s^{0}_1, S_2 = s^{*}_2) \nonumber\\
\nonumber\\
V_{1,2} \equiv F(S_1 = s^{*}_1, S_2 = s^{0}_2) - F(S_1 = s^{0}_1, S_2 = s^{0}_2) \nonumber
\end{eqnarray}
where $F$ is the characteristic/ value function
\newline
{\textbf{Benefits from increasing business 2’s surrogate value from $s^{0}_2$ to $s^{*}_2$:}}
\begin{eqnarray}
V_{2,1} \equiv F(S_1 = s^{*}_1, S_2 = s^{*}_2) - F(S_1 = s^{*}_1, S_2 = s^{0}_2) \nonumber\\
\nonumber\\
V_{2,2} \equiv F(S_1 = s^{0}_1, S_2 = s^{*}_2) - F(S_1 = s^{0}_1, S_2 = s^{0}_2) \nonumber
\end{eqnarray}

The attribution formula for this 2-business example will be the average of two possible cases above and is specified by
\begin{eqnarray}
\phi_1 = \dfrac{1}{2} \left(\dfrac{V_{1,1}}{1} + \dfrac{V_{1,2}}{1}\right)
\nonumber \\
\phi_2 = \dfrac{1}{2} \left(\dfrac{V_{2,1}}{1} + \dfrac{V_{2,2}}{1}\right) \nonumber
\end{eqnarray}

This generates a summary of the model using the baseline and realized levels $s^0_k$ and $s^*_k$ for each business in this case. Put differently, Shapley attributions are averages of these marginal contributions $\phi_k$ for each input $k$. Since there are many possible combinations of businesses being active (i.e., part of the coalition) or inactive, Shapley attributions serve as a way to compress these many potential outcomes into a single value. If
the Shapley attribution value for a business is close to zero it indicates that its average returns are small irrespective of other inputs.



\end{appendices}


\begin{thebibliography}{999}
\bibliographystyle{plain}

\bibitem{CausalML}
Chen, H., Harinen, T., Lee, J. Y., Yung, M., and Zhao, Z. (2020). CausalML: Python package for causal machine learning. arXiv preprint arXiv:2002.11631.

\bibitem{PyWhy1}
Sharma, A. and Kiciman, E., 2020. DoWhy: An end-to-end library for causal inference. arXiv
preprint arXiv:2011.04216.

\bibitem{PyWhy2}
Kiciman, E., Dillon E.W., Edge, D, Foster, A., Jennings, J., Ma, C., Ness, R., Pawlowski N., Sharma, A. and Zhang C., (2022) A Causal AI Suite for Decision-Making, NeurIPS 2022 Workshop on Causality for Real-world Impact

\bibitem{EconML1}
Syrgkanis, V., Lewis, G., Oprescu, M., Hei, M., Battocchi, K., Dillon, E., Pan, J., Wu, Y., Lo, P., Chen, H. and Harinen, T., (2021) Causal inference and machine learning in practice with EconML and CausalML: Industrial use cases at Microsoft, TripAdvisor, Uber. In Proceedings of the 27th ACM SIGKDD conference on knowledge discovery \& data mining (pp. 4072-4073).

\bibitem{EconML2}
Hartford, J., Lewis, G., Leyton-Brown, K., and Taddy, M. (2017). Deep IV: A flexible approach for counterfactual prediction. In International Conference on Machine Learning (pp. 1414- 1423). PMLR.

\bibitem{EconML3}
Athey, S., and Wager, S. (2021). Policy learning with observational data. Econometrica, 89(1), 133-161.

\bibitem{Pew_research_center} Quarterly Retail E-Commerce Sales Report, U.S. Census Bureau.



\bibitem{Tidio} Beginner’s Guide to Virtual Shopping Assistants \& Bots (2023) \href{https://www.tidio.com/blog/virtual-shopping-assistant/}{www.tidio.com/blog/virtual-shopping-assistant/}

\bibitem{AB} Young, Scott W. H. (2014). "Improving Library User Experience with A/B Testing: Principles and Process". Weave: Journal of Library User Experience. 1 (1). doi:\href{https://quod.lib.umich.edu/w/weave/12535642.0001.101?view=text;rgn=main}{10.3998/weave.12535642.0001.101}


\bibitem{More_DML} More, S., Kotwal, P., Chappidi, S., Mandalapu, D., Khawand, C. (2023). Double Machine Learning at Scale to Predict Causal Impact of Customer Actions.
Machine Learning and Knowledge Discovery in Databases: Applied Data Science and Demo Track. ECML PKDD 2023. Lecture Notes in Computer Science, vol 14174. Springer, Cham. \href{https://doi.org/10.1007/978-3-031-43427-3_31}{https://doi.org/10.1007/978-3-031-43427-3\_31} 

\bibitem{Spark} Apache Spark - \href{https://spark.apache.org/}{https://spark.apache.org/}.

\bibitem{Var1}
Athanasopoulos, G., Poskitt, D. S., and Vahid, F. (2012). Two canonical VARMA forms: Scalar component models vis-à-vis the echelon form. Econometric Reviews, 31(1), 60–83

\bibitem{Var2}
Hamilton, J. D. (1994). Time series analysis. Princeton University Press, Princeton.

\bibitem{Var3}
Lütkepohl, H. (2005). New introduction to multiple time series analysis. Berlin: Springer-Verlag.

\bibitem{Var4}
Lütkepohl, H. (2007). General-to-specific or specific-to-general modelling? An opinion on current econometric terminology. Journal of Econometrics, 136(1), 234–319.

\bibitem{Shapley1} Sundararajan, M. \& Najmi, A. (2019) The many Shapley values for model explanation. arXiv preprint arXiv:1908.08474.

\bibitem{Shapley2} Shapley, L.S. (1953) A value for n-person games. Contributions to the Theory of Games 2.28: 307-317.

\bibitem{AWSlambda} AWSlambda \href{https://aws.amazon.com/lambda/}{aws.amazon.com/lambda/}

\bibitem{mlflow} MLflow \href{https://mlflow.org/}{mlflow.org/}

\bibitem{SparkML} SparkML \href{https://spark.apache.org/docs/1.2.2/ml-guide.html}{spark.apache.org/ml-guide.html}

\bibitem{FPCI} Sekhon, Jasjeet. (2007) The Neyman–Rubin Model of Causal Inference and Estimation
via Matching Methods. The Oxford Handbook of Political Methodology.

\bibitem{attention} Vaswani, A., Shazeer, N., Parmar, N., Uszkoreit, J., Jones,
L., Gomez, A. N., Kaiser, Ł., and Polosukhin, I. (2017) Attention is all you need. In Advances in Neural Information
Processing Systems.

\end{thebibliography}
\end{document}